\documentclass[]{style/ceurart}
\usepackage{listings}
\usepackage{hyperref}
\usepackage{booktabs}

\begin{document}

\copyrightyear{2026}
\copyrightclause{Copyright for this paper by its authors.
  Use permitted under Creative Commons License Attribution 4.0
  International (CC BY 4.0).}

\conference{CLEF 2026 Working Notes, 21 -- 24 September 2026, Jena, Germany}

\title{Predicting Viticulture Potential through an Ensemble of U-Net and a Geospatial Foundation Model}

\subtitle{ImageCLEF AI4Agri Subtask 1 at CLEF 2026}

\author[1]{Jorge Ignacio Perez}[
    orcid=0009-0001-3367-7646,
    email=jperez333@gatech.edu,
]
\cormark[1]

\author[1]{Hwaai Kang Kee}[
    orcid=0009-0003-7613-5905,
    email=hkee7@gatech.edu,
]
\cormark[1]

\author[1]{Lucas Rassbach}[
    orcid=0009-0008-8096-7700,
    email=lrassbach3@gatech.edu,
]
\cormark[1]

\address[1]{Georgia Institute of Technology, North Ave NW, Atlanta, GA 30332}
\cortext[1]{Corresponding author.}

\begin{abstract}
    Determining agricultural potential is fundamental to sustainable land management and agricultural planning. Remote sensing data is increasingly valuable as an avenue for agricultural potential due to the cost of traditional methods (surveys, in-situ measurements, soil testing, etc). ImageCLEF AI4Agri 2026: Subtask 1 is concerned with the prediction of viticulture potential in Southern France. The DS@GT ARC's submission for Subtask 1 introduces an ensemble of U-Net and a Geospatial Foundation Model (Prithvi-2.0). Our best model achieved a $\pm$1 accuracy of 68.32 on the leaderboard, ranking 2nd among 7 teams. The implementation for this work is publicly available at \href{https://github.com/dsgt-arc/imageclef-ai4agri-2026}{github.com/dsgt-arc/imageclef-ai4agri-2026}.
\end{abstract}

\begin{keywords}
  ImageCLEF 2026 \sep
  AI4Agri 2026 \sep
  remote sensing \sep
  Agricultural Potential \sep
  Earth Observation \sep
  Precision Agriculture \sep
  Multi-temporal Imagery \sep
  Hyperspectral Data \sep
  Semantic Segmentation \sep
  Foundation Models \sep
  Vision Transformers \sep
  ViT
\end{keywords}

\maketitle

\section{Introduction}
The agricultural suitability of land is an important topic for agricultural entities such as farmers and policymakers, as well as for economic development efforts. If the agricultural potential of land is understood, a determination can be made as to what kind of agriculture should be facilitated, leading to better ecological conditions, higher crop yield, and improved economic outcomes. Agricultural suitability is typically done via physical surveying, a costly, time-consuming, and manual process~\cite{el2025agripotential}. A compounding disadvantage of the time-consuming and manual process of physical surveys is that the surveys must be reassessed frequently to capture the changing conditions of a region, especially when one considers climate change. The gains to be realized from the creation of a digital and automated process are profound. Subtask 1 of the ImageCLEF AI4Agri 2026 challenge~\cite{ionescu2026imageclef} utilizes the AgriPotential dataset~\cite{el2025agripotential} to classify Sentinel-2 satellite multispectral imagery, where each pixel is labeled with an agricultural suitability score from 1 to 5 for viticulture. Exploratory data analysis (EDA) indicates that temporal modeling appears to be important due to varying Normalized Difference Vegetation Index (NDVI) values across time frames. We hypothesize that temporal modeling is important and will lead to greater performance gains in this task.

The Data Science at Georgia Tech Applied Research Competitions (DS@GT ARC) team has developed a solution via an ensemble machine learning model consisting of a U-Net~\cite{ronneberger2015u} semantic segmentation model and the Prithvi~\cite{szwarcman2025prithvi} geospatial foundation model. Prithvi acts as the student in a teacher-student model, with the U-Net as the teacher for unlabeled pixels. This approach led to acceptable results and demonstrates additional avenues that could be pursued to improve performance further.

\section{Related Work}
\subsection{Prithvi-EO-2.0}
The paper by Szwarcman et al details an Earthly Observation (EO) pre-trained model, Prithvi-EO-2.0~\cite{szwarcman2025prithvi}. EO methods have been revolutionized by the introduction of AI systems trained on large unlabeled satellite imagery datasets. The authors created a new foundational model for EO machine learning tasks with the purpose of creating a general purpose baseline model. Prithvi-EO-2.0 is trained on 4.2 billion global time series samples with temporal and location embeddings. The model consists of the following architecture: a masked autoencoder (MAE) with a vision transformer (ViT) backend that encodes and decodes masked images. The model's performance was evaluated partially based on it's ability to classify several different types of land use, such as Natural Vegetation, Forest, Corn Agriculture, Developed, Ocean, and more, in the United States and Europe, across several datasets. For all datasets used to evaluate classification in Szwarcman et al, Prithvi-EO-2.0 achieved the best performance when compared against the predecessor Prithvi-EO-1.0 and a baseline U-Net model.

\subsection{U-Net}
Ronneberger et al discuss the U-Net model~\cite{ronneberger2015u}, a neural network methodology that is able to benefit from data augmentation for more efficient sample utilization when working with small image datasets. The network model architecture expands on convolutional networks by supplementing them with successive layers, with maxpools replaced by upsampling operators. These layers result in an increased resolution in the output. Additionally, the network has a large number of feature channels, which enable higher resolution layers in the network to have context information. The U-Net model results in good results on small training datasets when data augmentation is used, and is effective in the segmentation of images. When evaluated in the use of labeling per-pixel cell images and boundaries, the model outperformed other models in the ISBI cell tracking challenge in medical imagery.

The U-Net was used as a baseline in the AgriPotential paper~\cite{el2025agripotential}, which led us to experiment with it for this task.

\section{Methodology}
Our final submission consists of a weighted ensemble of two models: a U-Net trained on the full multi-temporal input by stacking all the timesteps and spectral bands as channels, and a Prithvi-EO-2.0 foundation model finetuned on seasonal aggregated inputs.
Both models are trained independently using the same ordinal loss formulation and are combined at inference time through a weighted logit ensemble, where the logits are linearly combined as $\mathbf{z} = w_{\text{U-Net}} \cdot \mathbf{z}_{\text{U-Net}} + w_{\text{Prithvi}} \cdot \mathbf{z}_{\text{Prithvi}}$, with weights $w_{\text{U-Net}} = 0.65$ and $w_{\text{Prithvi}} = 0.35$, calibrated empirically on the validation set.

\subsection{Dataset}
The AgriPotential dataset is introduced by El Sakka et al~\cite{el2025agripotential}, and is the basis of the ImageCLEF AI4Agri 2026: Subtask 1 challenge. The dataset is a collection of 34 temporal frames of multispectral Sentinel-2 satellite images from Southern France, with dates ranging from 2017 to 2019. Each pixel is labeled with potential for viticulture, market gardening, and field crops agriculture types, ranked from 1-5, with 1 being low in agricultural potential and 5 being high potential for agriculture. For the purpose of this challenge, we are only concerned with viticulture potential.

The size of the dataset is approximately 200GB. The dataset is organized into 6329 training patches, 781 validation patches, and 800 test patches. Each patch is a 128x128 pixel subset of an image containing a time series of the 34 time steps, 10 spectral bands, and pixel annotations of agricultural potential for the train and validation subsets. The dataset contains a high number of unlabeled pixels, with an average of 54\% of pixels labeled in each patch for training, 60\% labeled in validation, and 52\% labeled in test.

\subsection{Data augmentation}
\subsubsection{Spectral Indices}
\begin{figure}[htbp]
  \centering
  \includegraphics[width=\linewidth]{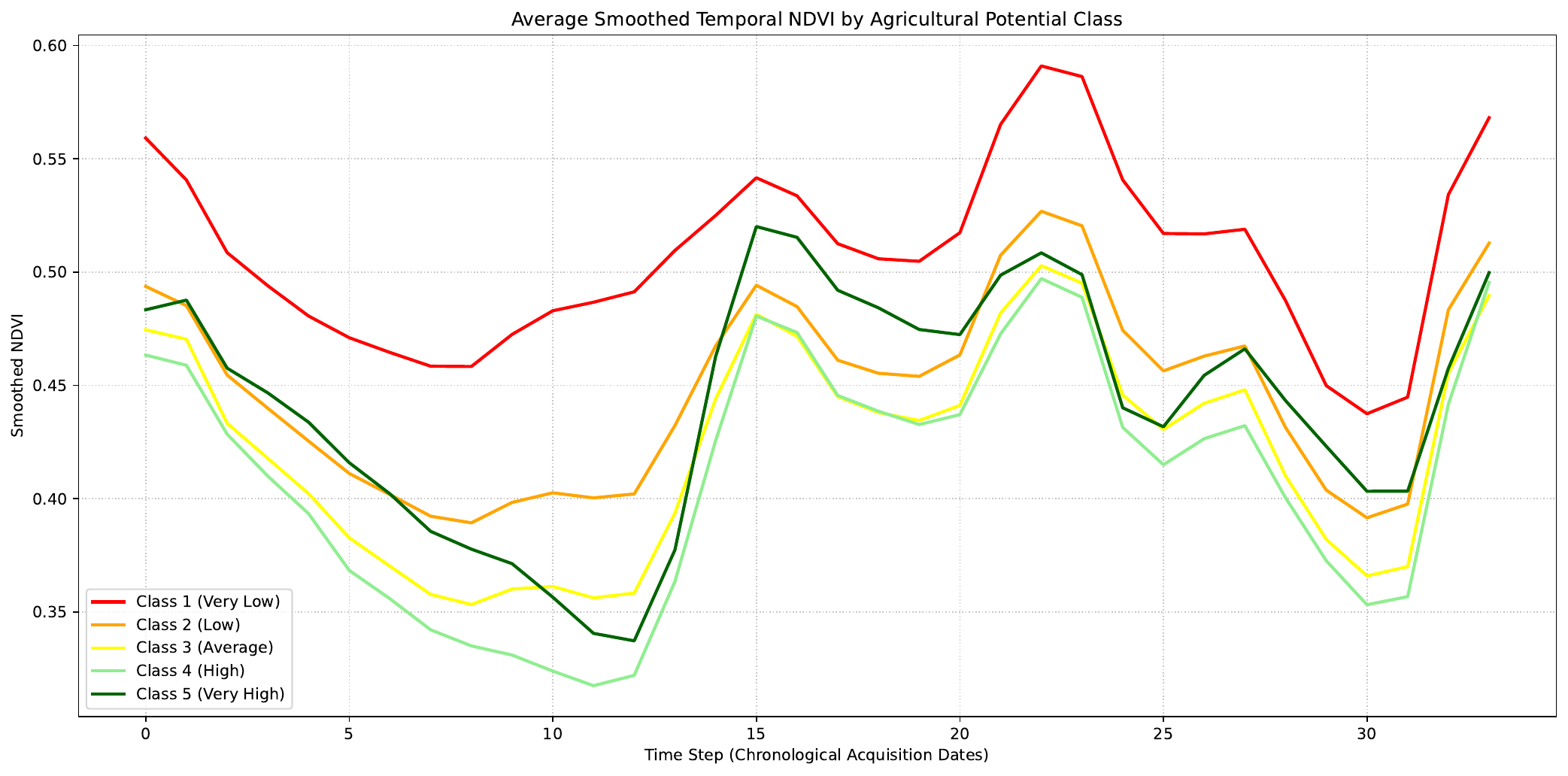}
  \caption{NDVI vs time steps for class labels}
  \label{fig:ndvi_class}
\end{figure}

NDVI is an index that provides an indication of vegetation status, providing an instrumental measure of environmental conditions, vegetation health, and vegetation productivity~\cite{rs16071212}. NDVI relies on multispectral data, lending itself well to be used in exploring the AgriPotential dataset. The NDVI index is plotted against the time step for each class in Figure~\ref{fig:ndvi_class}. An important observation of this analysis is that there is clear separation between the classes, with each class having a distinct NDVI range based on the timestep. An additional observation is that the NDVI index changes over time, with each class following the same general pattern. This indicates a temporal relationship.
The Normalized Difference Moisture Index (NDMI), Normalized Difference Water index (NDWI), Normalized Difference Red Edge (NDRE), and Normalized Burn Ratio (NBR) are not used for EDA but are used as auxiliary input channels in the model to provide complementary information on vegetation, moisture, and water.

\subsubsection{Image}
Rotation and flips were applied to the image in order to bridge the generalization gap between validation and test. Rotations and flips will reduce the model's reliance on local structure, thereby improving model performance in cases where a train-test distribution shift is present. Empirical experimentation results show that this improved model performance for both models in the ensemble.

\subsubsection{Normalization}
Three different normalization methods were tested: dividing by 10,000, Z-score normalization for the entire dataset split, and Z-score normalization per patch. For U-Net, experimentation demonstrated that dividing by 10,000 performed better than other methods. For Prithvi, Z-score normalization for the entire dataset split performed the best, maintaining consistency with the pretraining distribution. The final ensemble uses the best normalization method for its respective models.

\subsection{Model}
\subsubsection{U-Net}
Our first model is a U-Net architecture, motivated by the results presented by the AgriPotential paper~\cite{el2025agripotential}. In addition to the 10 raw Sentinel-2 spectral bands, we included five spectral indices as auxiliary input channels, resulting in a total of 15 channels. Temporal modeling is handled by stacking the $T=34$ timesteps and $C=15$ channels into a single input of $T \times C = 510$ channels, treating each timestep as additional input channels. Our model extends the standard U-Net with residual connections, and has a base channel dimension of 128 and depth 3. The dataset publication evaluated multiple loss formulations and found binary cross-entropy ordinal loss to outperform standard cross-entropy and mean squared error~\cite{el2025agripotential}; we confirm this finding in our own experiments and therefore adopt this loss to account for the ordered structure of the potential classes. Unlabeled pixels are excluded from the loss calculation so that only labeled regions contribute to training. The model produces a single-channel output that is passed through an ordinal layer composed of $K-1=4$ learnable thresholds. These are computed as the cumulative sum of softplus-transformed values, which guarantees positive and strictly increasing thresholds. Final logits are obtained by subtracting each threshold from the output, producing binary decision boundaries. U-Net architecture is summarized in Tables~\ref{tab:unet_architecture} and~\ref{tab:convblock_architecture}.

\begin{table}[htbp]
\centering
\begin{tabular}{l l l}
\toprule
\textbf{Layer} & \textbf{Description} & \textbf{Output Shape} \\
\midrule
\multicolumn{3}{l}{\textit{\textbf{Input}}} \\
Flatten & T=34 x C=15 & $(B, 510, 128, 128)$ \\
\midrule
\multicolumn{3}{l}{\textit{\textbf{Encoder}}} \\
ConvBlock + MaxPool2d & In: 510, Out: 128 & $(B, 128, 64, 64)$ \\
ConvBlock + MaxPool2d & In: 128, Out: 256 & $(B, 256, 32, 32)$ \\
ConvBlock + MaxPool2d & In: 256, Out: 512 & $(B, 512, 16, 16)$ \\
\midrule
\multicolumn{3}{l}{\textit{\textbf{Bottleneck}}} \\
ConvBlock & In: 512, Out: 1024 & $(B, 1024, 16, 16)$ \\
\midrule
\multicolumn{3}{l}{\textit{\textbf{Decoder}}} \\
Upsample + ConvBlock & In: 1024, Out: 512 & $(B, 512, 32, 32)$ \\
Upsample + ConvBlock & In: 512, Out: 256 & $(B, 256, 64, 64)$ \\
Upsample + ConvBlock & In: 256, Out: 128 & $(B, 128, 128, 128)$ \\
\midrule
\multicolumn{3}{l}{\textit{\textbf{Head}}} \\
Conv2d & Kernel: 1, In: 128, Out: 1 & $(B, 1, 128, 128)$ \\
\midrule
\multicolumn{3}{l}{\textit{\textbf{Ordinal layer}}} \\
Threshold subtraction & 4 learnable thresholds, cumsum(softplus(x)) & $(B, 4, 128, 128)$ \\
\bottomrule
\end{tabular}
\caption{U-Net model architecture.}
\label{tab:unet_architecture}
\end{table}

\begin{table}[htbp]
\centering
\begin{tabular}{l l}
\toprule
\textbf{Layer} & \textbf{Description} \\
\midrule
Conv2d & Kernel: 3, Padding: 1 \\
BatchNorm2d & $-$ \\
ReLU & Non-linear activation  \\
Dropout2d & p=0.2  \\
Conv2d & Kernel: 3, Padding: 1 \\
BatchNorm2d & $-$  \\
Residual Connection &  $-$\\
ReLU & Non-linear activation \\
\bottomrule
\end{tabular}
\caption{ConvBlock architecture. All layers preserve the input dimensions and output $(B, C_{out}, H, W)$.}
\label{tab:convblock_architecture}
\end{table}

The model was trained using the AdamW optimizer with a cosine annealing learning rate schedule, for a maximum of 50 epochs with early stopping patience of 10 based on validation loss. We also applied gradient clipping to stabilize training, particularly given the large input dimensionality. The reported hyperparameters were selected through empirical tuning on the validation set. The final configuration is summarized in Table~\ref{tab:unet_hyperparams}.

\begin{table}[htbp]
\centering
\begin{tabular}{l l}
\toprule
\textbf{Hyperparameter} & \textbf{Value} \\
\midrule
\multicolumn{2}{l}{\textit{\textbf{Architecture}}} \\
Base dim & 128 \\
Depth & 3 \\
Dropout & 0.2 \\
\midrule
\multicolumn{2}{l}{\textit{\textbf{Training}}} \\
Optimizer & AdamW \\
Weight decay & $1 \times 10^{-2}$ \\
Learning rate & $1 \times 10^{-4}$ \\
LR schedule & CosineAnnealingLR \\
Batch size & 32 \\
Max epochs & 50 \\
Early stopping patience & 10 \\
Gradient clipping & max\_norm=1.0 \\
\bottomrule
\end{tabular}
\caption{U-Net model hyperparameters.}
\label{tab:unet_hyperparams}
\end{table}

\subsubsection{Transformer-based approach}
With the goal of exploiting temporal dynamics in the data, we explored several transformer-based architectures, incorporating explicit temporal modeling, namely TSViT~\cite{tarasiou2023vits}, U-TAE~\cite{garnot2021panoptic}, Swin-V2~\cite{liu2022swin}, Presto~\cite{tseng2023lightweight}, and Prithvi-EO-2.0. Among these, the Prithvi model achieved the best individual results and therefore was included in the ensemble. We finetune the \textit{prithvi\_eo\_v2\_100\_tl} backbone (100M parameters) with a feature pyramid neck~\cite{lin2017feature} and a UperNet~\cite{xiao2018unified} decoder with 128 channels, followed by a regression head and the same ordinal layer as the U-Net model. The ordinal thresholds are initialized from the trained U-Net values rather than randomly, preserving the learned class boundaries and providing a more stable convergence while finetuning. Although larger Prithvi variants were also evaluated, they increased the train-validation gap without improving test performance, suggesting overfitting given the dataset size. Model architecture is detailed in Table~\ref{tab:prithvi_architecture}.

\begin{table}[htbp]
\centering
\begin{tabular}{l >{\raggedright\arraybackslash}p{6cm} l}
\toprule
\textbf{Layer} & \textbf{Description} & \textbf{Output Shape} \\
\midrule
\multicolumn{3}{l}{\textit{\textbf{Input}}} \\
Seasonal aggregation & 34 timestamps aggregated into 4 seasonal means (only 6 bands) & $(B, 6, 4, 128, 128)$ \\
\midrule
\multicolumn{3}{l}{\textit{\textbf{Backbone}}} \\
Prithvi-EO-2.0-100M & Pretrained ViT, temporal + location encoding & $(B, 257, 768)$ \\
\midrule
\multicolumn{3}{l}{\textit{\textbf{Neck}}} \\
Feature Pyramid Neck & Converts transformer outputs into hierarchical inputs for the decoder & $-$ \\
\midrule
\multicolumn{3}{l}{\textit{\textbf{Decoder}}} \\
UperNet & In: 256, Out: 128 & $(B, 128, 32, 32)$ \\
\midrule
\multicolumn{3}{l}{\textit{\textbf{Head}}} \\
Conv2d + BatchNorm2d + ReLU & Kernel: 3, Padding: 1, In: 128, Out: 128 & $(B, 128, 32, 32)$ \\
Dropout2d & p=0.1 & $(B, 128, 32, 32)$ \\
Conv2d & Kernel: 1, In: 128, Out: 1 & $(B, 1, 32, 32)$ \\
Upsample & Bilinear interpolation to input resolution & $(B, 1, 128, 128)$ \\
\midrule
\multicolumn{3}{l}{\textit{\textbf{Ordinal layer}}} \\
Threshold subtraction & 4 learnable thresholds, cumsum(softplus(x)) & $(B, 4, 128, 128)$ \\
\bottomrule
\end{tabular}
\caption{Prithvi finetuned model architecture.}
\label{tab:prithvi_architecture}
\end{table}

Due to computational constraints, we aggregated the 34 input timesteps into four seasonal means. This introduces an inductive bias toward seasonal agricultural patterns while remaining consistent with Prithvi-EO-2.0 pretraining configuration, facilitating transfer of pretrained features. To further match the pretraining setup, only the six spectral bands used during pretraining were included: Blue, Green, Red, NIR, SWIR-1, and SWIR-2. While this reduces the spectral input compared to the U-Net, it creates a different input representation which might help to capture complementary patterns across models, improving performance when combined in the ensemble.

Additionally, the model incorporates temporal and location coordinate encodings, as used during pretraining. Since inputs are seasonal aggregates rather than individual acquisition dates, we used fixed representative dates for each season. Geographic coordinates for each patch were extracted from the dataset metadata and provided to the model as location encodings.

During fine-tuning, the encoder is frozen for the first 7 epochs to allow the decoder and head to adapt before full end-to-end training. The model is trained using AdamW with a linear warmup of 3 epochs followed by a cosine annealing schedule, for a maximum of 40 epochs, with an early stopping patience of 10. Different learning rates were used, with a lower rate for the encoder than the decoder and head. Complete training hyperparameters are summarized in Table~\ref{tab:prithvi_hyperparams}.

\begin{table}[htbp]
\centering
\begin{tabular}{l l}
\toprule
\textbf{Hyperparameter} & \textbf{Value} \\
\midrule
\multicolumn{2}{l}{\textit{\textbf{Architecture}}} \\
Backbone & prithvi\_eo\_v2\_100\_tl \\
Decoder channels & 128 \\
Head dropout & 0.1 \\
\midrule
\multicolumn{2}{l}{\textit{\textbf{Finetuning}}} \\
Optimizer & AdamW \\
Weight decay & $1 \times 10^{-2}$ \\
Encoder LR & $1 \times 10^{-5}$ \\
Decoder/head LR & $5 \times 10^{-5}$ \\
LR schedule & Linear warmup + CosineAnnealingLR \\
Warmup epochs & 3 \\
Backbone frozen epochs & 7 \\
Batch size & 32 \\
Max epochs & 40 \\
Early stopping patience & 10 \\
Precision & bf16-mixed \\
\midrule
\multicolumn{2}{l}{\textit{\textbf{Semi-supervised}}} \\
Teacher confidence threshold & 0.7 \\
Pseudo-label weight $\lambda$ & 0.3 \\
\bottomrule
\end{tabular}
\caption{Prithvi finetuned model hyperparameters.}
\label{tab:prithvi_hyperparams}
\end{table}

As mentioned previously, the dataset contains patches with unlabeled pixels, which are excluded from the loss calculations. To make use of this data, we adopt a simplified teacher-student pseudo-labeling approach. The trained U-Net functions as a fixed teacher, generating pseudo-labels for unlabeled pixels when the prediction confidence exceeds a threshold of 0.7. These labels are incorporated into the student training loss with weight $\lambda=0.3$, providing an additional training signal that helps regularize training and encourages smoother decision boundaries.

\section{Results}
\begin{table}[htbp]
\centering
\begin{tabular}{l c c c}
\toprule
\textbf{Model} & \textbf{Val $\pm$1 Accuracy} & \textbf{Val Exact Accuracy} & \textbf{Test $\pm$1 Accuracy} \\
\midrule
Ensemble (U-Net + Prithvi) & 0.8026 & 0.4360 & \textbf{68.32} \\
U-Net & 0.8083 & 0.4233 & 66.25 \\
Prithvi & 0.7593 & 0.4075 & 65.51 \\
TSViT & 0.81 & - & 64.0 \\
U-TAE & 0.81 & - & 57 \\
Swin-V2 & 0.822 & - & 54.92 \\
Presto & 0.78 & - & 49 \\
\bottomrule
\end{tabular}
\caption{Metrics of evaluated models. Test metrics are obtained from the public leaderboard.}
\label{tab:model_metrics}
\end{table}

Final results for all models on the validation and test sets are shown in Table~\ref{tab:model_metrics}. Our final submission achieved a test $\pm$1 accuracy of 68.32\%, placing 2nd on the competition leaderboard. The ensemble outperformed both individual models, with the U-Net and Prithvi models achieving 66.25\% and 65.51\%, respectively. We also include the results from earlier experiments with the different transformer architectures presented in the methodology section. 

\begin{figure}[htbp]
  \centering
  \includegraphics[width=\linewidth]{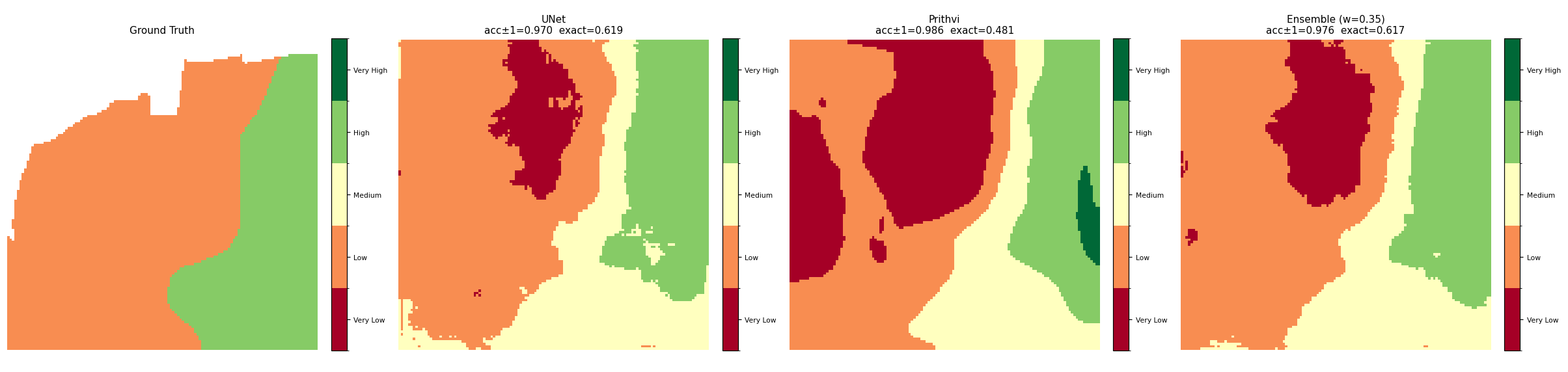}
  \caption{Comparison of ground truth and predictions on a patch with high validation accuracy across models.}
  \label{fig:prediction_comparison_1}
\end{figure}

\begin{figure}[htbp]
  \centering
  \includegraphics[width=\linewidth]{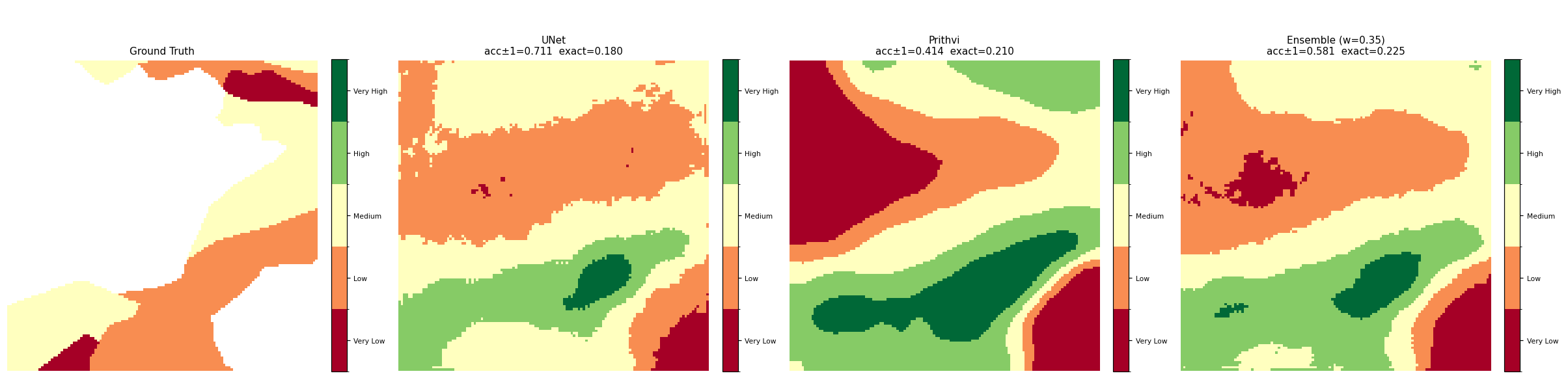}
  \caption{Comparison of ground truth and predictions on a more challenging patch with substantially lower accuracy.}
  \label{fig:prediction_comparison_2}
\end{figure}

To complement the quantitative results, Figures~\ref{fig:prediction_comparison_1} and~\ref{fig:prediction_comparison_2} present prediction maps comparing the U-Net, Prithvi, and ensemble outputs on two representative validation patches. In Figure~\ref{fig:prediction_comparison_1}, all models achieve strong performance on a relatively clean patch where class regions are well separated and only a small localized area is unlabeled. The U-Net produces noisier predictions, while Prithvi generates smoother outputs, more consistent spatially with the ground truth. 

Figure~\ref{fig:prediction_comparison_2} shows a different and more challenging scenario. Both models struggle in regions with multiple neighboring classes and complex boundaries, which is further complicated by larger unlabeled gaps. As seen previously, Prithvi still produces more spatially coherent predictions, but some regions are consistently misclassified.

\section{Discussion}
\subsection{Generalization gap}
As shown in Table~\ref{tab:model_metrics}, there is a generalization gap between validation and test $\pm$1 accuracy across all models, particularly among the earlier transformer-based models. Prithvi generalized more effectively in comparison, which can in part be attributed to the strong alignment between the pretraining setup and the target task, as well as the teacher-student approach which may have provided a slight regularizing effect by incorporating unlabeled pixels during training. However, this was still a consistent challenge throughout the task. We partially addressed it using the ensemble along with image augmentation techniques such as flips and rotations, both of which improved model performance and which is why they are applied to the final model. A train-test distribution shift is suspected, but further work is needed to make this claim. Class distribution on the test set should be examined, along with the percentage of unlabeled pixels, to determine its sparseness. Investigation on the geographical regions for the test set could determine if the terrain is different between the train and test sets (e.g.: mountainous regions vs flat lands).

\subsection{Temporal modeling}
Even though initial EDA indicated that temporal modeling is useful due to varying NDVI values across time frames as shown in Figure~\ref{fig:ndvi_class}, empirical experimentation with various model backbones with temporal modeling, such as Prithvi, TSViT~\cite{tarasiou2023vits}, U-TAE~\cite{garnot2021panoptic}, Swin-V2~\cite{liu2022swin}, and Presto~\cite{tseng2023lightweight}, did not yield better test $\pm$1 accuracy as shown in Table~\ref{tab:model_metrics}. Results suggest that it is more useful to view the time frames as additional channels due to the fixed time frames (2017-2019) between train/validation/test. Semantic segmentation is performed across time frames instead of for each time frame. This is why the U-Net with 510 stacked channels performed better than other models with temporal modeling. By stacking the time frames as channels, an inductive bias is introduced to the model where it views different time frames as distinct observations to the same pixel, which leads to better performance since we are performing semantic segmentation per patch instead of across different time frames. This contradicts our initial hypothesis that temporal modeling will lead to better model performance in this task.

However, it is important to note that Prithvi with seasonal aggregation performed reasonably well with only a difference of 0.74 in $\pm$1 test accuracy. For Prithvi-EO-2.0, the time frames are determined through a sampling method with some restrictions~\cite{szwarcman2025prithvi}. Sequences of 4 images are sampled through the Harmonized Landsat and Sentinel-2 (HLS) dataset~\cite{ju2025harmonized} to provide sufficient representation of seasonal changes. Consecutive images in a sequence are chosen with a minimum interval of 1 month and a maximum of 6 months between them. Even though no seasonal aggregation was performed during the pretraining of Prithvi-EO-2.0, we find that the sampling method used to select time frames for sequences of images during the pretraining of Prithvi-EO-2.0 translates well to our seasonal aggregation method. Further work is needed to determine if explicit temporal modeling is useful for this dataset.

\subsection{Ensemble}
The best performing model was the ensemble between U-Net and Prithvi as shown in Table~\ref{tab:model_metrics}. The ensemble improved results by helping two weak models generalize better. U-Net learns more fine-grained spatial patterns, while Prithvi might be capturing seasonal patterns due to the aggregated seasonal inputs. As shown in Figure~\ref{fig:prediction_comparison_1} for patches with high validation accuracy, the ensemble combines the characteristics from the two, preserving the overall predicted classes while reducing local prediction noise. Interestingly, both models share similar difficulties near class boundaries, which seem to be the main source of exact prediction errors. While the ensemble improved overall performance, the improvement was not consistent across all patches. As observed in Figure~\ref{fig:prediction_comparison_2}, it can inherit errors from both models, sometimes leading to worse predictions than an individual model.

\section{Future Work}
\subsection{Generalization gap}
Future work is needed to determine the cause of the generalization gap between validation and test. Right now, it is unclear if the generalization gap is caused by the model overfitting to training data or if a distribution shift occurred between the validation and test sets. It would be useful to examine the class distribution on the test set, along with the percentage of unlabeled pixels, as well as the means and standard deviations of the reflectance values.

\subsection{Larger Prithvi and full temporal frame}
Due to hardware constraints, seasonal aggregation was used instead of the full 34 time frames. Even though Prithvi-EO-2.0-300M did not lead to better performance in our case, compared to Prithvi-EO-2.0-100M. It is worthwhile to try Prithvi-EO-2.0-300M with the full 34 time frames to see if performance improves.

\subsection{Exclusion of non-informative data}
Certain images have a high percentage of unlabeled data. In such images, the loss calculation for these data is uninformative, and it might be harmful to include these images in training data. Excluding these data might improve model performance. This might be an interesting avenue to experiment with.

\section{Conclusions}
This work presents a practical approach for predicting viticulture potential from multitemporal Sentinel-2 imagery. Our method combines a U-Net which treats temporal timesteps as stacked input channels, with a Prithvi-EO-2.0 foundation model finetuned on seasonal aggregates to leverage its large-scale geospatial pretraining. The final ensemble achieved a $\pm1$ test accuracy of 68.32, placing 2nd on the public leaderboard.

Our results indicate that for this task, explicit temporal modeling provided limited improvements, while the convolutional approach remained a strong baseline compared to more complex temporal architectures. The generalization gap was a consistent challenge across all models, which we partially mitigated through ensembling, data augmentation and semi-supervised training. However, fully understanding the source of these generalization issues is still an open direction for future work.

\section*{Acknowledgements}

We thank the Data Science at Georgia Tech (DS@GT) CLEF competition group for their support.
This research was supported in part through research cyberinfrastructure resources and services provided by the Partnership for an Advanced Computing Environment (PACE) at the Georgia Institute of Technology, Atlanta, Georgia, USA~\cite{PACE}. 

\section*{Declaration on Generative AI}
During the preparation of this work, the authors used Grammarly in order to: Grammar and spelling check. The authors also used Gemini and Claude in order to: Sentence structure and phrasing. After using these tools, the authors reviewed and edited the content as needed and take full responsibility for the publication’s content.

\bibliography{main}
\end{document}